# Neural Networks in Adversarial Setting and Ill-Conditioned Weight Space

Mayank Singh[*1], Abhishek Sinha[*1], and Balaji Krishnamurthy[1]

[1]Adobe Systems Inc, Noida, India

*Abstract*— Recently, Neural networks have seen a huge surge in its adoption due to their ability to provide high accuracy on various tasks. On the other hand, the existence of adversarial examples have raised suspicions regarding the generalization capabilities of neural networks. In this work, we focus on the weight matrix learnt by the neural networks and hypothesize that ill conditioned weight matrix is one of the contributing factors in neural network's susceptibility towards adversarial examples. For ensuring that the learnt weight matrix's condition number remains sufficiently low, we suggest using orthogonal regularizer. We show that this indeed helps in increasing the adversarial accuracy on MNIST and F-MNIST datasets.

## I. INTRODUCTION

Deep learning models have performed remarkably well in several domains such as computer vision [20], [21], [22], natural language processing [24], [25] and speech recognition [23]. These models are able to achieve high accuracy in various tasks and hence their recent popularity. Due to their adoption in diverse fields, the robustness and security of Deep Neural Networks becomes a major issue. For the reliable application of Deep Neural Networks in the domain of security, the robustness against adversarial attacks must be well established. In recent work, it was shown that Deep Neural Networks are highly vulnerable to adversarial attacks [15]. The adversarial attacks are hand-crafted inputs on which the neural network behaves abnormally. Generally, in these kind of attacks a small magnitude of calculated noise is added to an input instance of training data to make the model output a significantly different result had it been on the unaltered input instance. In the case of images, some of the perturbations are so subtle that the adversarial and original training images are humanly indistinguishable. The existence of adversarial examples compels one to think about the generalization and learning capabilities of neural networks.

There had been several speculative explanation regarding the existence of adversarial examples. Some of the explanations attribute this to the non-linearity of deep neural networks, but recently in [14] the authors showed that linear behavior in high dimensional spaces is sufficient to produce adversarial examples in neural networks. Our work further builds upon this explanation by performing this linear computation of neural networks in high dimension close to a well conditioned space for increased stability against malicious perturbations.

*These authors contributed equally

## II. RELATED WORK

Various adversarial attacks and protection methods have been proposed in the existing literature. Some of the well-known attacks are the Fast Gradient Sign Method (FGSM) [14], Basic Iterative Method (BIM) [4], RAND+FGSM [17], DeepFool [9], Black-Box Attack [4], [5], Jacobian-Based Saliency Map Attack [16] and the L-BFGS Attack [15].

We are briefly going to describe some of the attacks that were used in our experiments. In a neural network, let $\theta$ denote its parameters, $x$ be the input to the model from the domain $[0, 1]^d$, $y$ be the true output label/value for input $x$ and $J(\theta, x, y)$ be the cost function.

### A. Fast Gradient Sign Method

In the FGSM attack [14] the adversarial example is constructed by using:

$$x^{adv} = x + \epsilon sign(\nabla_x J(\theta, x, y))$$

Here, $x^{adv}$ is the adversarial example generated using input $x$ and $\epsilon$ is the variable reflecting the magnitude of perturbation that is being introduced while constructing the adversarial example. Some of the adversarial images generated from MNIST dataset using this attack for different $\epsilon$ values are shown in Fig 1.

### B. Basic Iterative Method

The BIM [4] is an extension of FGSM where adversarial examples are crafted by applying FGSM multiple times with small step size($\alpha$). Clipping of pixel values of intermediate results is done to ensure that each pixel perturbation magnitude doesn't exceed $\epsilon$. Here, n denotes the number of iterations to be applied.

$$x_0^{adv} = x$$

$$x_{n+1}^{adv} = Clip_{x,\epsilon}\{x_n^{adv} + \alpha sign(\nabla_x J(\theta, x_n^{adv}, y))\}$$

### C. RAND+FGSM

The RAND+FGSM [17] is a modification of FGSM where the FGSM is applied on the data point $x^{'}$ which is obtained by adding a small random perturbation of step size $\alpha$ to the original data point $x$.

$$x^{'} = x + \alpha sign(\mathcal{N}((0^d, I^d)))$$

$$x^{adv} = x^{'} + (\epsilon - \alpha)sign(\nabla_{x^{'}} J(\theta, x^{'}, y))$$

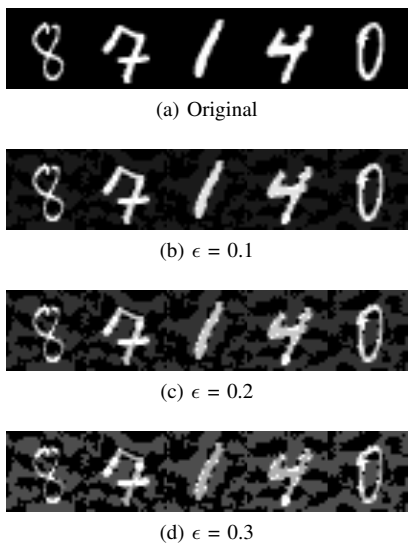

(a) Original

(b) $\epsilon = 0.1$

(c) $\epsilon = 0.2$

(d) $\epsilon = 0.3$

Fig. 1: (a) Original test sample images which the network correctly classifies. (b), (c) ,(d) represents a sample of the corresponding adversarial images generated via FGSM for different $\epsilon$ values. For $\epsilon$ taking values 0.1, 0.2 and 0.3, the model mis-classifies 1, 4 and 5 of the above images respectively.

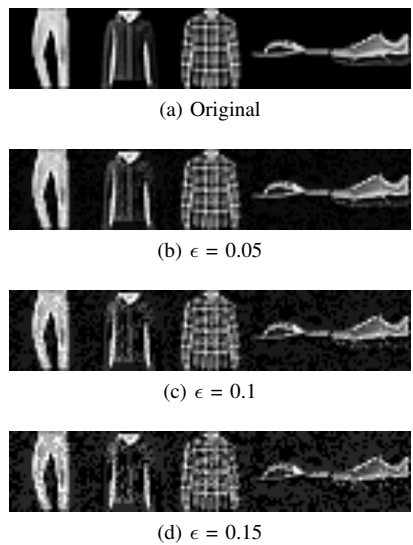

(a) Original

(b) $\epsilon = 0.05$

(c) $\epsilon = 0.1$

(d) $\epsilon = 0.15$

Fig. 2: (a) Original test sample images which the network correctly classifies. (b), (c) ,(d) represent a sample of the corresponding adversarial images for different $\epsilon$ values generated via the black box attack.

*D. Practical Black-Box Attack*

Black-box attacks [4], [5] doesn't require any prior information regarding the structure of architecture or the parameters learnt by the target model. As the name suggests only the labels corresponding to inputs are required to construct the adversarial examples. These attacks are based on the premise of transferability of adversarial examples between different architectures of deep neural network trained on the same data [14], [15]. One of the black-box attacks [5] comprises of training a local substitute model to simulate the target model. In this approach it is assumed that the attacker has a small set of inputs which were drawn from the same input distribution as that of the training data used for the target model. The training data of substitute model consists of synthetically generated data using the given small set of inputs. The labels for this training data is obtained by querying the target model. A sample of adversarial generated examples of F-Mnist dataset for different $\epsilon$ values are shown in Fig 2.

As the types of adversarial attacks are increasing in number, so are the defensive techniques to protect deep learning models. Yet there is no general defense mechanism which guarantees robustness against all of the existing attacks. Some of these adversarial defense techniques include ideas such as training on adversarial examples [14], using ensemble models [1], [17], adding entropy regularizer on the output probabilities [6] and distillation [7].

The property of orthogonality has a huge appeal in mathematical operations due to the inherent stability that comes with it. Random orthogonal initial condition on weight matrices in neural networks has been shown to retain finite learning speed even in case of deep neural architecture [18]. Furthermore, retaining this orthogonal property of weight matrix have helped in fixing the problem of exploding and vanishing gradients, particularly in case of Recurrent Neural Networks enabling them to learn long term dependencies [12], [10], [11]. In order to learn rich internal representation autoencoders have been used with a regularizer that encourage the hidden layers to learn orthogonal representation of input [2]. In domain adaptation techniques, some improvements were made by learning dissimilar private and shared representation. It was obtained by enforcing soft orthogonality optimization constraints on the private and shared representations [3]. Therefore, orthogonality constraints have been used for an array of tasks which span from learning rich representation in latent space to fixing the problem of exploding and vanishing gradients. We will see that it also has utility in facilitating reduction of condition number of neural network's weight space in adversarial setting.

## III. THEORY

Condition number of a matrix or linear system [26] measures the sensitivity of the matrix's operation in the event of introducing perturbation to inputs or the resulting value. Condition number is a norm dependent property and in this paper we will focusing on 2-norm. Orthogonal matrix have a condition number of 1 whereas singular matrix have infinitely large condition number.

Matrices that have condition number close to that 1 are said to be "well-conditioned" and those which are close to singular matrix(i.e. have large condition number) are said to be "ill-conditioned". Condition number of a matrix is also representative of its sensitivity in inverse computation

The condition number of a Matrix A is defined as:

$$\kappa(A) = ||A||.||A^{-1}|| \quad (1)$$

where norm of the matrix is defined by

$$||A|| = \max_{x \neq 0} \frac{||Ax||}{||x||}$$

Consider a system of linear equation.

$$Ax = b \quad (2)$$

The effect of the perturbation in x and b can be described by using condition number of A.

$$A(x + \delta x) = b + \delta b \quad (3)$$

$$\frac{||\delta x||}{||x||} \leq \kappa(A) \frac{||\delta b||}{||b||} \quad (4)$$

We can use this analysis to consider the case of a fully connected layer in neural network. As the intermediate computations consists of linear equations:

$$Wx + b = p \quad (5)$$

where $W$ is the weight matrix, $b$ are the biases, $x$ is the input signal and $p$ is the output before passing it through activation layer.

Combining b and p to get :

$$Wx = b_1 => W^{-1}b_1 = x \quad (6)$$

As the condition number a matrix and its inverse is the same, given any perturbation in $x$, $b_1$, using (2) and (4) we can write :

$$\frac{||\delta b_1||}{||b_1||} \leq \kappa(W) \frac{||\delta x||}{||x||} \quad (7)$$

As adversarial examples are malicious perturbation added to the input ($x$) of the model, improving the condition number of the weight space ($\kappa(W)$) limits the changes in the intermediate-output ($b_1$), which can seen from (7).

Similarly, this can be extended to convolutional neural networks by focusing on the condition number of the matrix formed where each row denotes the filter's weight optimized by the neural network. For example, in a particular convolutional neural network with parameters $(K_x, K_y, C_{in}, N_f)$ where

- $K_x$ - is the x-dimension of the filter
- $K_y$ - is the y-dimension of the filter
- $C_{in}$ - is the number of input channels of the input image
- $N_f$ - is the number of filters used in the model

One can visual these learnable parameters of network as a matrix having dimension $((K_x \times K_y \times C_{in}), N_f)$ and carry out the same analysis as done earlier in the case of fully connected layer.

## IV. PROPOSED SOLUTION

As we have seen in the previous section, condition number of the weight matrix can play an important role in deciding the amount of change observed in the intermediate layer's output while dealing with perturbed input. Hence an effort in reducing the condition number of weight space of the neural network should consequently increase neural network's robustness in adversarial setting. To achieve our goal of pushing the weights towards well-conditioned space, we propose using orthogonal regularizer as a heuristic inspired from the fact that orthogonal matrices have the ideal condition number. While training we propose adding an extra loss reflecting the penalty for ill-conditioned weight denoted by

$$L_{cond} = \lambda(W^T.W - I)$$

Here $W$ is the l2-normalized weight matrix for a particular layer of neural network, $\lambda$ is the condition loss regularization hyperparameter and $I$ is the identity matrix of suitable dimension. So for a classification task the total loss to be optimized becomes:

$$L_{total} = L_{classification} + L_{cond}$$

This $L_{cond}$ is different for each layer and can be applied over all the layers of neural network with different settings of $\lambda$ as required.

## V. EXPERIMENTS AND RESULTS

To understand the effectiveness of our approach we consider two different types of adversarial attacks that are used for neural networks :-

- White box attacks - Here the attacker has the complete knowledge of the model architecture that was used for training as well as the data with which the model was trained. The attacker can then use the same model architecture and training data to train the model and then generate adversarial examples .
- Black box attacks - Here the attacker has no knowledge of the model architecture used to train for the desired task. It also does not have access to the data used for training. To generate adversarial examples the attacker thus needs to train a substitute network and generate its own data. The attacker can however query the actual model to get the labels corresponding to the generated training set.

We evaluated our approach on the FGSM, RAND+FGSM and BIM white box attacks as well as FGSM black box attack. In order to verify if our approach can be applied along with approaches that aim to minimize the risk of adversarial attack, we applied our method on adversarial training and evaluated the results.

We conducted all our experiments on two different datasets : the MNIST handwritten dataset and the Fashion-MNIST clothing related dataset[19]. Both the datasets consist of 60,000 training images and 10,000 test images. The images are gray-scale of size 28×28. For all the white box experiments we generated adversarial examples from the test set consisting of 10k images.

*A. Results on white box attacks*

In this section we present results on white box attacks using three different methods :- FGSM, RAND+FGSM and BIM.

*1) FGSM attack:* We tested our approach on the following two neural network architectures:-

- A convolutional neural network(A) with 2 convolutional layers and 2 fully connected layers(dropout layer after 1st fc layer is also present) with ReLU activations. Max pooling(2×2 pool size and a 2×2 stride) was applied after every conv layer. The CNN layer weights were of shape [5, 5, 1, 32] and [5, 5, 32, 64] respectively and the fc layer were of sizes [3136, 1024] and [1024, 10].

- A fully connected neural network(B) with 2 hidden units each consisting of 256 hidden units and ReLU activation.

We also trained the network using adversarial training($\epsilon = 0.3$) and further applied our approach on top of it to check if our method can be used on top of other methods for preventing adversarial attack or not.

The regularization parameter($\lambda$) used in our approach for each of the different layers was selected on the basis of observing the condition number of each layer by observing orthogonal regularization loss during training. Layers having higher condition numbers were assigned larger values of $\lambda$ compared to those having low condition numbers. We stress here that the hyperparameter $\lambda$ was chosen not on the basis of the adversarial accuracy of the model on the test set but rather on the basis of condition numbers of layers and the validation set classification accuracy. We need to take into consideration the validation set classification accuracy because larger values of $\lambda$ lead to reduction in accuracy.

We tested the FGSM attack over MNIST dataset for different values of $\epsilon$ and the results are shown in tables I and III for the two network architectures. As can be inferred from the results our approach improves the adversarial accuracy under both the cases :- when directly applied as a regularizer and when applied as a regularizer over the adversarial training approach. The second result is interesting because it suggests the possibility of further improvement when our method is augmented with other techniques that have been proposed to improve adversarial accuracy. We have not shown the performance of network B for high values of $\epsilon$ because the performance of the network becomes already very bad even at $\epsilon = 1.5$ for adversarial examples.

Similar experiments were performed over the F-MNIST dataset for the two different network architectures and the results have been shown in tables II and IV. We see that under normal training the adversarial accuracy drops very low for high values of $\epsilon$ and our approach also does not improve the accuracy under these settings.

TABLE I: Adversarial accuracy for FGSM attack over MNIST dataset for Network A

| $\epsilon$ | Normal | Regz. | Adv. tr. | Adv. tr.+Regz. |
|---|---|---|---|---|
| 0.05 | 0.9486 | **0.9643** | 0.9752 | **0.9768** |
| 0.1 | 0.7912 | **0.8759** | 0.9527 | **0.9656** |
| 0.15 | 0.4804 | **0.6753** | 0.9352 | **0.9678** |
| 0.2 | 0.1903 | **0.3847** | 0.9212 | **0.9741** |
| 0.25 | 0.058 | **0.1484** | 0.9008 | **0.9787** |
| 0.3 | 0.0238 | **0.0276** | 0.8729 | **0.979** |

TABLE II: Adversarial accuracy for FGSM attack over F-MNIST dataset for Network A

| $\epsilon$ | Normal | Regz. | Adv. tr. | Adv. tr.+Regz. |
|---|---|---|---|---|
| 0.05 | 0.5013 | **0.5559** | **0.7728** | 0.7713 |
| 0.1 | 0.2128 | **0.274** | 0.6926 | **0.7073** |
| 0.15 | 0.0658 | **0.1007** | 0.6261 | **0.6535** |
| 0.2 | 0.01 | **0.0227** | 0.5564 | **0.5862** |
| 0.25 | **0.0026** | 0.0022 | 0.4763 | **0.5071** |
| 0.3 | **0.0004** | 0.0003 | 0.4153 | **0.4454** |

TABLE III: Adversarial accuracy for FGSM attack over MNIST dataset for Network B

| $\epsilon$ | Normal | Regz. | Adv. tr. | Adv. tr.+Regz. |
|---|---|---|---|---|
| 0.025 | 0.8895 | **0.9194** | 0.9387 | **0.9449** |
| 0.05 | 0.5819 | **0.7256** | 0.8345 | **0.8612** |
| 0.075 | 0.237 | **0.3872** | 0.6063 | **0.6903** |
| 0.1 | 0.0731 | **0.1603** | 0.3362 | **0.4446** |
| 0.125 | 0.032 | **0.0539** | 0.1527 | **0.2254** |
| 0.15 | **0.0198** | 0.017 | 0.0689 | **0.0998** |

TABLE IV: Adversarial accuracy for FGSM attack over F-MNIST dataset for Network B

| $\epsilon$ | Normal | Regz. | Adv. tr. | Adv. tr.+Regz. |
|---|---|---|---|---|
| 0.025 | 0.5459 | **0.5844** | **0.7592** | 0.7521 |
| 0.05 | 0.225 | **0.2928** | 0.5816 | **0.597** |
| 0.075 | 0.0787 | **0.1088** | 0.3994 | **0.4398** |
| 0.1 | 0.0295 | **0.0319** | 0.236 | **0.2875** |
| 0.125 | **0.0114** | 0.005 | 0.1285 | **0.1751** |
| 0.15 | **0.0041** | 0.0008 | 0.0613 | **0.0897** |

We have shown maximum of the condition number of different layers in network in table V. The condition number of the layers were calculated via the matrix 2 norm. As can be seen from the table, adding the loss corresponding to orthogonality of the weights does indeed reduce the condition number of the weight matrices.

To see how our approach affects the test accuracy of the network, we have shown the result in table VI. As can be seen from the table, our method does not much affect the test accuracy for both the two datasets. The same is true even when the approach is applied on top of adversarial training method. Thus we can say that our method does improve the adversarial performance of the networks without any compromise with the test accuracy.

TABLE V: Max condition number of network weights

| Dataset | Net | Normal | Regz. | Adv.tr. | Adv.tr.+Regz. |
|---------|-----|--------|-------|---------|---------------|
| MNIST   | A   | 17.56  | 3.73  | 121.78  | 23.49         |
|         | B   | 995.70 | 251.19| 1192.47 | 14.88         |
| F-MNIST | A   | 15.94  | 5.63  | 114.30  | 23.14         |
|         | B   | 513.33 | 49.01 | 875.87  | 26.48         |

TABLE VI: Test Accuracy of networks under different settings

| Dataset | Net | Normal | Regz. | Adv.tr. | Adv.tr.+Regz. |
|---------|-----|--------|-------|---------|---------------|
| MNIST   | A   | 0.9916 | 0.9916| 0.9917  | 0.9907        |
|         | B   | 0.9777 | 0.9789| 0.9803  | 0.979         |
| F-MNIST | A   | 0.9038 | 0.9016| 0.8892  | 0.8852        |
|         | B   | 0.8898 | 0.8847| 0.8841  | 0.8814        |

*2) RAND+FGSM and BIM attack:* For the RAND+FGSM attack a Gaussian noise was added to the examples before subjecting them to the FGSM attack. The value of $\alpha$ was kept to be 0.5 and experiments were conducted for the two datasets for different $\epsilon$ values. The results have been shown in table VII.

TABLE VII: Adversarial accuracy for RAND+FGSM for Network A

| Dataset | $\epsilon$ | Normal | Regz. |
|---------|------------|--------|-------|
| MNIST   | 0.05       | 0.9911 | **0.9915** |
|         | 0.1        | 0.9411 | **0.9587** |
|         | 0.15       | 0.7582 | **0.8536** |
|         | 0.2        | 0.4171 | **0.6183** |
|         | 0.25       | 0.1333 | **0.3186** |
|         | 0.3        | 0.0379 | **0.0983** |
| F-MNIST | 0.05       | **0.896** | 0.8944 |
|         | 0.1        | 0.4686 | **0.5223** |
|         | 0.15       | 0.1879 | **0.2417** |
|         | 0.2        | 0.05   | **0.0805** |
|         | 0.25       | 0.0065 | **00.0135** |
|         | 0.3        | **0.0017** | 0.0008 |

For the BIM attack $\alpha$ was kept to be 0.025 and the value of $n$ was 2, 3, 6, 9 corresponding to the different $\epsilon$ values. The results for the experiment have been shown in table VIII. The results show that our method makes the network be robust to all the three different types of adversarial attack without affecting the test accuracy performance of network.

### B. Results on black box attacks

For the black box attack we created a substitute network with the following architecture:-
A fully connected neural network(C) with 2 hidden units each consisting of 200 hidden units and ReLU activation.

The substitute network had access to only 150 test samples initially and new data was augmented to it for $n = 6$ times via the Jacobian based data augmentation technique. Network A was used as the classifier for this attack. Adversarial examples were generated using the trained substitute network which were then subsequently fed for classification to the original classifier.

The results over the generated adversarial samples are shown in table IX for the two datasets MNIST and F-MNIST. As can be seen from the results, our approach does improve the performance of the network over adversarial examples generated from the substitute network across different values of $\epsilon$ for both the datasets.

TABLE VIII: Adversarial accuracy for BIM for Network A

| Dataset | $\epsilon$ | Normal | Regz. |
|---------|------------|--------|-------|
| MNIST   | 0.025      | 0.9433 | **0.9622** |
|         | 0.05       | 0.8575 | **0.9173** |
|         | 0.1        | 0.2047 | **0.4635** |
|         | 0.15       | 0.007  | **0.0322** |
| F-MNIST | 0.025      | 0.4737 | **0.5287** |
|         | 0.05       | 0.2816 | **0.343**  |
|         | 0.1        | 0.0172 | **0.0306** |
|         | 0.15       | 0      | **0.0001** |

TABLE IX: Adversarial accuracy under Black Box attack

| Dataset | $\epsilon$ | Normal | Regz. |
|---------|------------|--------|-------|
| MNIST   | 0.05       | 0.9879 | **0.9887** |
|         | 0.1        | 0.9817 | **0.984**  |
|         | 0.15       | 0.9686 | **0.9765** |
|         | 0.2        | 0.9481 | **0.9624** |
|         | 0.25       | 0.9076 | **0.9359** |
|         | 0.3        | 0.8256 | **0.8752** |
| F-MNIST | 0.05       | 0.8565 | **0.8667** |
|         | 0.1        | 0.7858 | **0.8161** |
|         | 0.15       | 0.6924 | **0.7456** |
|         | 0.2        | 0.577  | **0.6453** |
|         | 0.25       | 0.459  | **0.5328** |
|         | 0.3        | 0.3505 | **0.4319** |

## VI. DISCUSSION

In the previous section we showed results as to how reducing the condition number of weight matrices via forcing them to align orthogonally helped in performance over adversarial examples. In this section we try to see some other issues that a network could face as a result of high condition number in the it's layers.

The condition number of a matrix in the case of 2-norm becomes the ratio of largest to smallest singular value. Consider a square matrix $A$ of n dimension having the singular value decomposition (SVD) [27] as $A = U\Sigma V^T$. Rewriting the SVD of $A$ as a combination of n equations where $i \in \{1, 2, .., n\}$ we have:

$$Av_i = \sigma_i u_i \quad (8)$$

$$\kappa(A) = \frac{\sigma_1}{\sigma_n} \quad (9)$$

If the matrix is ill-conditioned then one of the following is the case: either $\sigma_1$ is high or $\sigma_n$ is low or both. From (8) and (9), we can observe by perturbing the input in the direction of $v_n$ and applying it to $A$ produces least amount of change in output. In fact $v_n$ forms the least square solution of $Ax = 0$. Hence, in an ill-conditioned weight matrix of neural network with sufficiently low value of $\sigma_n$, perturbing the input in the direction of right singular vector $v_n$ will produce minimum change magnitude-wise when applied over matrix.

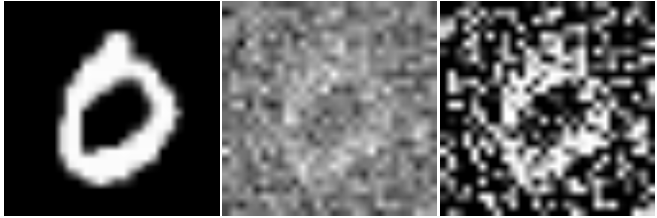

(a) Original    (b) Unclipped perturbed    (c) Clipped perturbed

Fig. 3: (a) Original test sample image of class 0 which the network correctly classifies with high confidence of 0.999 (b) represents the unclipped perturbed test sample image in the direction of minimum eigenvector with $\lambda = 20$ as mentioned in (10). The confidence of classification for class 0 for the original and regularized classifiers were 0.999 and 0.105 respectively.(c) represents the clipped(between 0 and 1) perturbed test sample image generated with the same configuration as that of (b). For (c) the confidence of classification for class 0 in case the of original and regularized classifier were 0.916 and 0.454 respectively.

$$A(x + \lambda v_n) = Ax + \lambda(\sigma_n u_n) \qquad (10)$$

Leveraging this observation in a fully connected neural network, we generated data points which were significantly different from the original data point taken from Mnist dataset by keeping a reasonably high value of $\lambda$. The model was still predicting it to be of the same class as that of the original data point. These artificially generated data points can be thought of as another type of adversarial examples which are visibly different from the original data points but the models labels them the same with high confidence. Examples of the generated examples along with their predictions can be seen in Fig 3. In the same architecture of fully connected network with condition number penalty applied, significant drop in the confidence of labelling was observed. Hence, we can say that more sensible results are generated when models are regularized while keeping condition number of the weight in check.

## VII. CONCLUSION AND FUTURE DIRECTION

In this paper we have explored the relationship between the condition number of the weights learnt by neural network and it's vulnerability towards adversarial examples. We have shown theoretically that well-conditioned weight space of neural networks are relativity less prone to be fooled by adversarial examples by the means of inferring bounds on change in output with respect to input in neural layers. We have validated our theory on various adversarial techniques and datasets. One of the heuristics that was used to control the condition number of weight space was orthogonal regularizer, but any other approach that influences the condition number in the positive light should also work. Application of the proposed technique should help in creating more robust neural networks specially in security related fields. In future work, we would like to explore adversarial generation techniques and feasibility of preconditioning in context of neural networks.